\documentclass{article}

\PassOptionsToPackage{square, numbers}{natbib}
\usepackage[final]{neurips_2025}

\makeatletter
\renewcommand{\@makefntext}[1]{%
  \noindent
  \makebox[1.2em][l]{\@makefnmark}%
  \footnotesize  
  #1}
\makeatother
\makeatletter
\renewcommand{\@makefntext}[1]{%
  \parindent 0em%
  \noindent
  \hb@xt@1.8em{\hss\@makefnmark}%
  \scriptsize
  #1}
\makeatother

\usepackage{hyperref}
\usepackage[utf8]{inputenc} 
\usepackage[T1]{fontenc}    
\usepackage{hyperref}       
\usepackage{url}            
\usepackage{graphicx}
\usepackage{booktabs}       
\usepackage{amsfonts}       
\usepackage{nicefrac}       
\usepackage{microtype}      
\usepackage{xcolor}         
\usepackage{enumitem}
\setlist[itemize]{leftmargin=8pt, labelindent=10pt, labelwidth=60em, labelsep=0.5em}
\usepackage{amsmath} 
\usepackage{amsfonts}       
\usepackage[table]{xcolor} 
\usepackage{multirow}

\usepackage[textsize=tiny]{todonotes}
\newcommand{\userprompt}[1]{\todo[color=purple!20, inline, author=USER]{#1}}

\usepackage{pifont} 
\usepackage{caption}
\title{Knowledge-based Visual Question Answer with Multimodal Processing, Retrieval and Filtering}

\author{
Yuyang Hong$^{1,2}\footnotemark[1]$, 
Jiaqi Gu$^{3}\footnotemark[1]$, 
Qi Yang$^{1,2}$, 
Lubin Fan$^{3}\footnotemark[2]$, 
\And
Yue Wu$^{3}$,
Ying Wang$^{2}$, 
Kun Ding$^{2}\footnotemark[2]$, 
Shiming Xiang$^{1,2}$, 
Jieping Ye$^{3}$\\
\\
$^1$School of Artificial Intelligence, University of Chinese Academy of Sciences\\
$^2$MAIS, Institute of Automation, Chinese Academy of Sciences\\
$^3$Alibaba Cloud Computing
}

\begin{document}
\footnotetext[1]{Equal contribution: \texttt{Yuyang Hong <hongyuyang2023@ia.ac.cn>, Jiaqi Gu <vadin@zju.edu.cn>}}
\footnotetext[2]{Corresponding authors: \texttt{Lubin Fan <lubin.flb@alibaba-inc.com>, Kun Ding <kun.ding@ia.ac.cn>}}

\maketitle
\vspace{-14pt}
\begin{abstract}
\vspace{-5pt}
Knowledge-based visual question answering (KB-VQA) requires visual language models (VLMs) to integrate visual understanding with external knowledge retrieval. Although retrieval-augmented generation (RAG) achieves significant advances in this task by combining knowledge-base querying, it still struggles with the quality of multimodal queries and the relevance of retrieved results.
To overcome these challenges, we propose a novel three-stage method, termed Wiki-PRF, including \textbf{P}rocessing, \textbf{R}etrieval and \textbf{F}iltering stages. The processing stage dynamically invokes visual tools to extract precise multimodal information for retrieval. The retrieval stage integrates visual and text features to achieve multimodal knowledge retrieval. 
The filtering stage performs relevance filtering and concentration on retrieval results.
To this end, we introduce a visual language model trained with answer accuracy and format consistency as reward signals via a reinforcement learning manner. This enhances the model's reasoning, tool invocation for accurate queries, and filtering of irrelevant content. Experiments on benchmark datasets (E-VQA and InfoSeek) show significant improvements~(36.0 and 42.8) in answer quality, achieving state-of-the-art performance. Code is available at: \url{https://github.com/cqu-student/Wiki-PRF}
\vspace{-8pt}
\begin{figure}[h]
    \centering
    \includegraphics[width=1.0\linewidth]{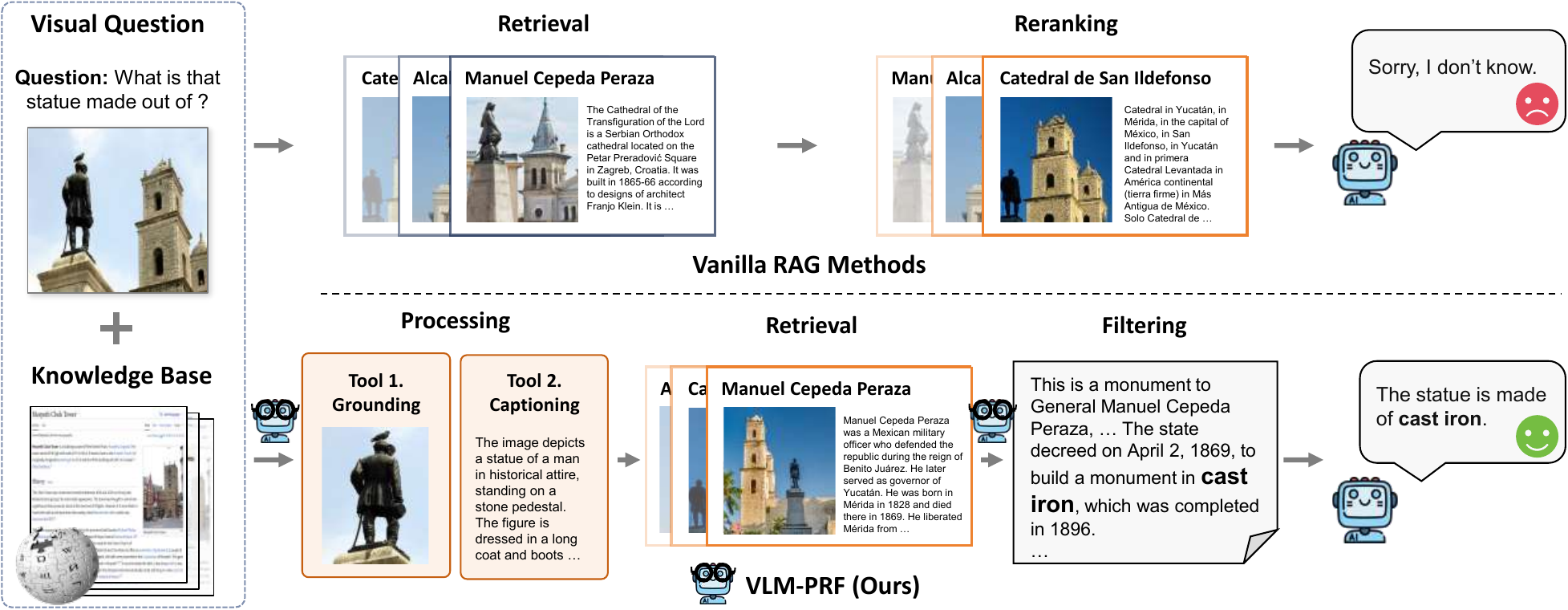}
    \caption{\textbf{Illustration of vanilla RAG methods and our Wiki-PRF.}
    Different from the traditional RAG methods (above), our method (below) employs multimodal tools processing stage and a further filtering stage, enabling more effective retrieval and extraction of task-relevant information.}
    \label{fig: guanggao}
    \vspace{-0.5cm}
\end{figure}
\end{abstract}

\section{Introduction}
Visual Language Models (VLMs)~\cite{achiam2023gpt,bai2025qwen2,zhu2025internvl3} have demonstrated remarkable capabilities in Visual Question Answering (VQA) tasks~\cite{antol2015vqa,goyal2017making}. Despite their effectiveness, they still face challenges in addressing knowledge-based visual question answering (KB-VQA), as such questions require not only an understanding of visual content but also the integration of external knowledge. For instance, answering the question "What is that statue made of?" in Figure \ref{fig: guanggao} requires factual knowledge that goes beyond the visual content.
To address this issue, retrieval-augmented generation (RAG) methods enhance model performance by incorporating mechanisms to access and integrate external information. These methods typically involve three steps: first, retrieving knowledge based on the given visual or textual content; second, reranking the retrieved information for relevance; and third, generating answers using the most pertinent content.

However, current methods~\cite{lin2022retrieval,gao2022transform,ding2022mukea,yan2024echosight,wu2022multi,chen2022murag} often fail to retrieve the most relevant information when handling complex visual content and questions, leading to suboptimal answers. This issue primarily arises from two key challenges. (1) Fine-grained knowledge retrieval in complex visual scenes: When process information-rich images, existing retrieval methods that rely on full-image visual features are often insufficiently precise for effective knowledge retrieval. For example, when asking about statues near a bell tower in Figure~\ref{fig: guanggao}, the statue may occupy only a small portion of the image. Consequently, the retrieval process can be heavily influenced by the more prominent bell tower, introducing excessive irrelevant information related to it.
(2) Precise filtering of irrelevant information from large scale retrieved results: After retrieving contextual information, it is difficult to filter out irrelevant or low-quality content using paragraph reranking alone. The retrieved results typically contain significant amounts of extraneous information, which can affect the accuracy of the generated answers.

To address these challenges, we propose a novel multimodal RAG method consisting of three stages: processing, retrieval, and filtering. The core idea is to obtain more relevant knowledge to generate accurate answers. For coarse retrieval limitations, we innovatively explored a tool-based fine-grained retrieval mechanism. For irrelevant information, we innovatively employ a question-based filtering stage. Specifically,
in the processing stage, the visual language model autonomically invokes image-processing tools based on the input image and question. These tools perform operations such as image captioning, visual grounding, and image flipping to extract detailed information related to the question from the image, thereby generating high-quality multimodal retrieval inputs.
In the retrieval stage, multimodal retrieval is conducted using both visual features and text descriptions to retrieve relevant knowledge.
In the filtering stage, the retrieved contextual information is filtered and condensed to remove redundancy, extract the most relevant knowledge, and provide to the answer generator for generating accurate responses.
To this end, we introduce Wiki-PRF, a RAG method that not only supports basic multimodal question-answering functionality but also enhances reasoning based on the input image and question. Wiki-PRF can flexibly invoke visual tools and demonstrates stronger capabilities in filtering and condensing retrieval results.

To enable the visual language model to possess the aforementioned reasoning ability, we train a VLM-PRF model using reinforcement learning~(RL). This is because training data collected for complex visual question-answering tasks often lacks the intermediate reasoning steps, which are necessary for effective supervised fine-tuning of VLMs.
RL~\cite{mnih2013playing,van2016deep,schaul2015prioritized}, as a paradigm for learning strategies to achieve specific goals, has been widely adopted in recent years to enhance the reasoning capabilities of VLMs~\cite{guo2025deepseek,zhou2025r1,meng2025mm,ahmadian2024back,kool2019buy,yang2025r1} for specialized tasks. RL can utilize a small amount of sample data, relying solely on answer accuracy as the reward signal, to train the model to generate high-quality retrieval content by accurately invoking task-specific tools. Additionally, it enables the model to selectively retain and condense the most relevant retrieval results for the query. 
Specifically, we employ the LoRA~\cite{hu2022lora} to train only a small number of additional parameters, enabling our Wiki-PRF to enhance its RAG capabilities without compromising its core question-answering abilities. 
In summary, our main contributions are as follows:
\begin{itemize}
    \item A knowledge-based visual question-answering method using a Processing-Retrieval-Filtering framework is proposed, named Wiki-PRF. It effectively leverages external tools for information retrieval and systematically filters the retrieved knowledge to support the generation of precise answers.
    \item We introduce VLM-PRF, a visual language model for multimodal RAG tasks, trained via reinforcement learning to enhance reasoning. To our knowledge, this represents the first application of reinforcement learning to multimodal retrieval-augmented generation, requiring minimal training data while enabling flexible tool use and robust processing.
    \item Comprehensive experiments demonstrate that Wiki-PRF achieves state-of-the-art performance on E-VQA~(36.0) and InfoSeek ~(42.8). Additional analyses further validate our method's effectiveness.
\end{itemize}

\section{Related Work}
\subsection{Knowledge-based Visual Question Answering}
Knowledge-Based Visual Question Answering (Knowledge-Based VQA)~\cite{deng2025comprehensive}
As a critical branch of Visual Question Answering (VQA), Knowledge-Based VQA demands models to integrate the understanding of visual content and question with external knowledge bases for reasoning and answering. Based on knowledge base modalities, Knowledge-Based VQA frameworks can be categorized into unimodal \cite{marino2019ok,schwenk2022okvqa,wang2017fvqa,lin2023fine,lin2022retrieval,gao2022transform} and multimodal \cite{mensink2023encyclopedic,chen2023can,ding2022mukea,yan2024echosight,wu2022multi,chen2022murag} paradigms. Uni-modal methods \cite{lin2023fine,lin2022retrieval,gao2022transform} typically utilize text-only datasets such as Wiki-21M\cite{karpukhin2020dense} and GS112K\cite{luo2021weakly} as external knowledge sources. For unimodal methods, TRiG\cite{gao2022transform} facilitates knowledge passage retrieval and generative question answering by converting images into plain text, thus fully harnessing the power of large-scale knowledge bases and pre-trained language models. \\
For multimodal methods \cite{ding2022mukea, yan2024echosight,wu2022multi,chen2022murag,cocchi2024augmenting}, external knowledge bases typically incorporate datasets such as Encyclopedic VQA (E-VQA)\cite{mensink2023encyclopedic} and InfoSeek\cite{chen2023can}, which include both Wikipedia images and corresponding textual information. MuKEA\cite{chen2022murag} represents multimodal knowledge through explicit triplets to capture the implicit relationships between visual objects and factual answers. EchoSight\cite{yan2024echosight} first retrieves candidate Wikipedia articles using visual information, then re-ranks them based on text-image query relevance to improve retrieval performance. Unlike previous methods, our method enhances the utilization of external knowledge bases by enabling the model to autonomously select and filter relevant information during the retrieval processing.

\subsection{Reinforcement Learning for Visual Language Model}
Reinforcement learning (RL) \cite{mnih2013playing,van2016deep,schaul2015prioritized}, a learning paradigm that improves model decision-making through interaction with an environment and feedback in the form of rewards, has recently been widely applied to vision-language models \cite{alayrac2022flamingo,bai2025qwen2,koh2023grounding,li2023blip,wang2024qwen2} (VLMs). Some works \cite{guo2025deepseek,zhou2025r1,meng2025mm,ahmadian2024back,kool2019buy,yang2025r1} focus on enhancing the reasoning capabilities of Vision-Language Models (VLMs) through reinforcement learning. R1-OneVision\cite{yang2025r1} framework innovatively bridges vision and language by encoding visual data into formalized textual representations, enabling robust and precise reasoning grounded in linguistic semantics. VisualThinker-R1-Zero\cite{zhou2025r1} achieves the first successful realization of an 'Aha Moment' in multimodal reasoning using a 2B-parameter VLM. Other works\cite{lumathvista,wang2024measuring,zhang2024mathverse,liu2025visual,shen2025vlm} focus on leveraging reinforcement learning to improve the performance of Vision-Language Models in specific areas like mathematical reasoning and visual perception. Visual-RFT~\cite{liu2025visual} leverages VLMs to generate reasoning-enhanced responses and integrates task-specific verifiable rewards (e.g., IoU for detection) via policy optimization methods like GRPO~\cite{shao2024deepseekmath}, improving model performance. In contrast to above work, our study uniquely introduces RAG capabilities into VLMs via RL. As far as we are aware, this is the first exploration of RL-based method for RAG in VLM.

\begin{figure}[t!]
    \centering
    \includegraphics[width=1\linewidth]{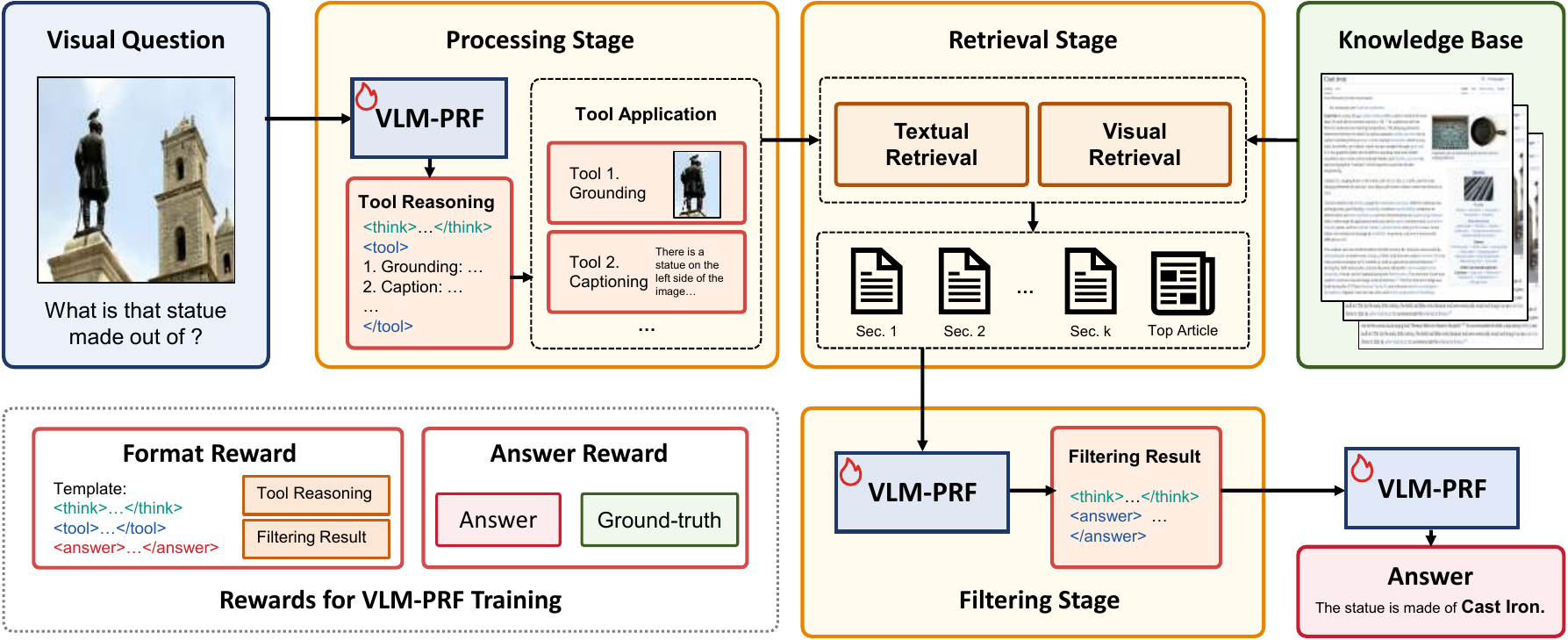}
   
    \caption{\textbf{Overview of Wiki-PRF}. 
    Wiki-PRF comprises three key stages:
    (1) Processing Stage: VLM-PRF processes the input image and its corresponding question using external tools. 
    (2) Retrieval Stage:  Relevant Wikipedia articles are retrieved, split into individual sections, and ranked based on their similarity to the processed input.
    (3) Filtering Stage: The re-ranked article sections are further filtered by VLM-PRF to retain the most relevant content, which is then fed into the VLM for final answer generation.
    During training, VLM-PRF is supervised using two types of reward signals: answer reward , which evaluates the correctness of the generated answer, and format reward , which ensures the output adheres to the desired structure.
    }
    \label{fig: main figure}
    \vspace{-0.5cm}
\end{figure}
\section{Method}

\subsection{Overview}
Knowledge-based VQA requires answering question $Q$ that is highly relevant to a given reference image $I$, with the assistance of a knowledge base KB. In our setup, KB $\in \{(a_{1}, I_{1}), . . . ,(a_{n}, I_{n})\}$ consists of a million-scale collection of entity articles $\{a_i\}$ along with their corresponding image set $\{I_i\}$. Our goal is to improve multimodal retrieval quality by flexibly invoking visual tools and enhancing relevance through filtering and enrichment.

As illustrated in Figure~\ref{fig: main figure}, the overall architecture of Wiki-PRF consists of three key components: an external knowledge base (KB), a model (VLM-PRF) trained via reinforcement learning, and a base model without extra trained parameters. 
The Wiki-PRF method comprises three main stages: (1) Processing Stage: The VLM-PRF model invokes external tools to process the raw reference image $I$ and question $Q$, generating precise retrieval queries $Query$. (2) Multimodal Retrieval Stage: The model performs multimodal information retrieval based on the generated query $Query$ and retrieves relevant information from the knowledge base. (3) Filtering Stage: The VLM-PRF model filters and extracts highly relevant information from the retrieval results and structures it into task-oriented knowledge, which is subsequently utilized to augment the answer.

\subsection{Processing Stage}

Previous methods~\cite{mensink2023encyclopedic,chen2023can,ding2022mukea,yan2024echosight,wu2022multi,chen2022murag} rely on raw input for retrieval, often missing key details due to a lack of interactive processing. For instance, a statue next to a church might be overlooked in favor of the church itself.
To address this, our method introduces tool-based preprocessing before retrieval, enhancing results through secondary data refinement. We employ several representative tools:
1) A captioning tool that captures high-level semantic information from images. 2) A grounding tool that extracts regions of interest for precise, detailed retrieval. 3) A flipping tool that adjusts the image’s orientation to mitigate the impact of direction on retrieval. 
Through these tools, Wiki-PRF achieves more comprehensive and accurate retrieval results. In essence, VLM-PRF provides the strategy while VLM-base delivers the core tool functionality, for captioning and grounding, which we define as VLM$_{\text{captioning}}$ and VLM$_{\text{grounding}}$, respectively.

\begin{figure}[t!]
    \centering
    \includegraphics[width=1\linewidth]{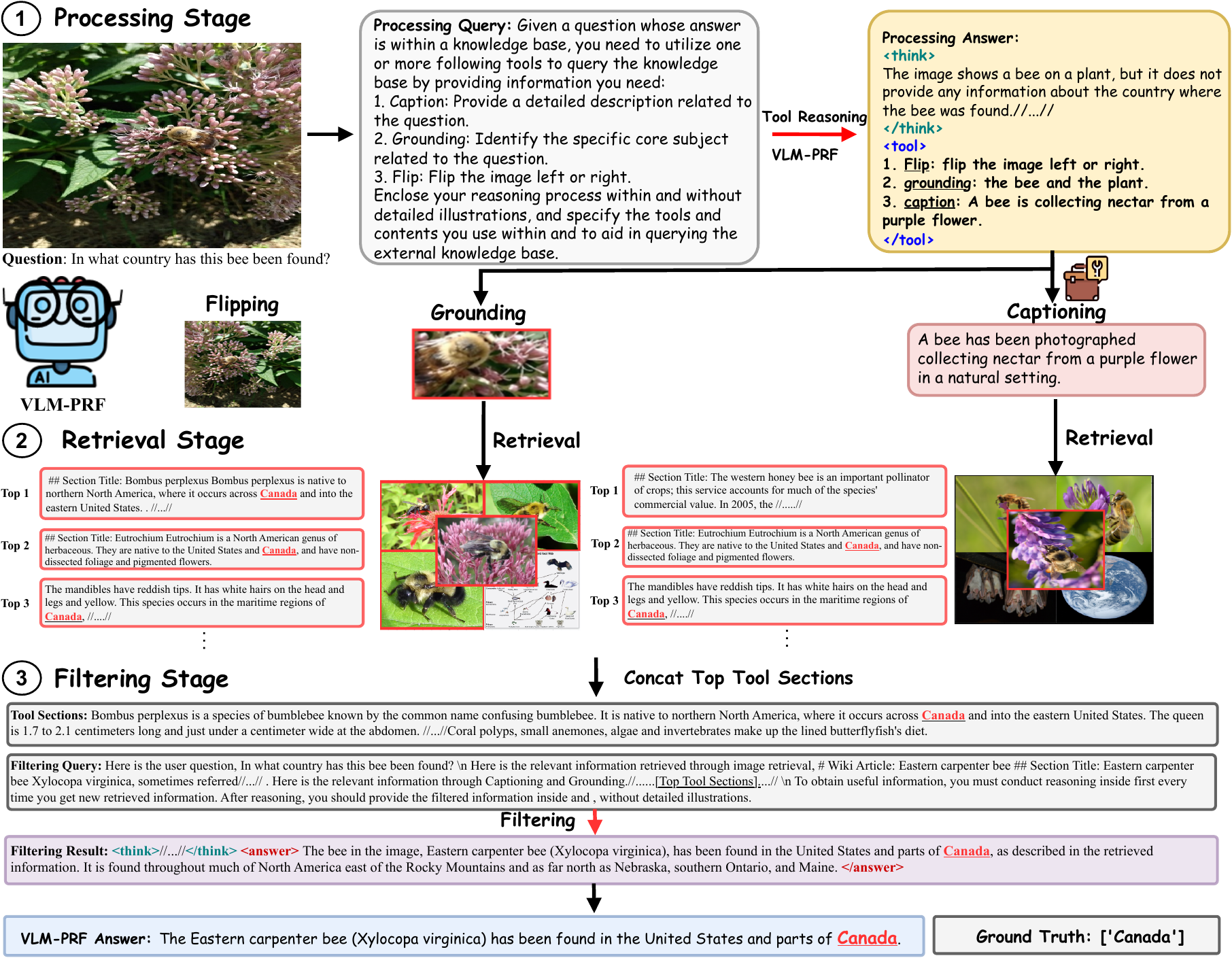}
    \caption{\textbf{Example of the tool calls and filtering}. By analyzing the problem, VLM-PRF performs captioning, grounding and flipping operations on the images. Using the retrieved sections, VLM-PRF performs filtering and generates the task-oriented results.}
    \label{fig:tool}
\end{figure}

As in Figure~\ref{fig:tool}, given an image $I$ and a question $Q$, the VLM-PRF model reasons about which tools to use and in what order within \textcolor{teal}{$<$think$>$} tags, then outputs selected tools and their execution order in \textcolor{blue}{$<$tool$>$} tags. After VLM-PRF plans the sequence of tool calls, the tasks are executed by VLM-base, a foundational model (Qwen2.5-VL-7B). This model is invoked multiple times to power specific tools like captioning and grounding. For the captioning tool, VLM$_{\text{captioning}}$ takes the init caption $C_{init}$ generated by VLM-PRF as input and produces the final caption $C_{query}$ for retrieval:
\begin{equation}
    C_{query} = \text{VLM}_{\text{captioning}}(C_{init},Q),
\end{equation}
Specifically, VLM-PRF first outputs the $C_{init}$ to be processed by the captioning tool $\text{VLM}_{\text{captioning}}$. Then $\text{VLM}_{\text{captioning}}$ employs $C_{init}$ as input to generate the final query $C_{query}$.

For grounding tool, VLM$_{\text{grounding}}$ takes the object output by VLM-PRF and returns the positional information. The image $I$ is then corpped based on positional information and generate $I_{\text{grounding}}$: 
\begin{equation}
    I_{\text{grounding}} = \text{Crop}(I,\text{VLM}_{\text{grounding}}(\text{object})).
\end{equation}
The flipping tool applies a left-right inversion to reference image in order to mitigate the impact of angular variations on the retrieval performance. Finally, all $Query$ results generated by the tools are aggregated to perform refined and accurate retrieval.


\subsection{Multimodal Retrieval Stage}
The goal of multimodal retrieval is to retrieve relevant articles $a$ from a knowledge database $KB$, based on the reference image $I$ and the generated $Query$. Initially, retrieval is performed directly on $I$, and the most relevant article obtained from this process is selected as the base retrieval information, denoted as $D$. Leveraging the $Query$ produced by the tools, additional information is retrieved from the knowledge base to enrich the search context. 

    

\paragraph{Tool Search.} The retrieval query $Query$ is first embedded into a feature vector using EVA-CLIP~\cite{sun2023eva}. We then employ the Faiss library~\cite{douze2024faiss} with cosine similarity $S^{D}_{\text{tool}}$ to efficiently retrieve the top-k most relevant images and their corresponding documents from the knowledge base:
\begin{equation}
    S^{D}_{\text{tool}} = \max_{k}\left\{ \left\langle \frac{\boldsymbol{T}}{\|\boldsymbol{T}\|} \cdot \frac{\boldsymbol{V}_i}{\|\boldsymbol{V}_i\|} \right\rangle, i = 1, \dots, n \right\},
\end{equation}
\begin{equation}
    (\mathcal{A},\mathcal{I})_{\text{tool}} = \{(\mathcal{A}_{i},\mathcal{I}_{i}),i\in S^{D}_{\text{tool}}\},
\end{equation}
\begin{equation}
    \boldsymbol{T} = \Phi(Query),     \boldsymbol{V}_i = \Phi(I_{i}),
\end{equation}

where $(\mathcal{A},\mathcal{I})$ denotes the top-$k_{D}$ retrieved articles and their associated images, $\Phi$ represents the feature extractor of EVA-CLIP, $\boldsymbol{T}$ is the embedding of the $Query =\{C_{\text{init}},I_{\text{grounding}}\}$, and $\boldsymbol{V}_i$ is the visual embedding of image $I$. The retrieved articles $\mathcal{A}$ are split into sections $\mathcal{S}$, and after removing titles, sections will be selected based on cosine similarity $S^{s}_{\text{tool}}$. For captioning, we calculate the cosine similarity between $Query$ and sections. For grounding, we directly calculate the cosine similarity between the question embedding,  $T'= \Phi (Q)$ and sections to maximize the fusion of modal information. The top-$k_{s}$ most similar sections are then selected as the final retrieval results, where $s^{j}_{i}$denotes the $j$-th section of the $i$-th article:
\begin{equation}
    \mathcal{A}_i = \{s^{i}_{1},s^{i}_{2},...,s^{i}_{m} \},i=1,\dots,k_{s},
\end{equation}
\begin{equation}
    S^{s}_{\text{tool}} = \max_{k_{s}} \left\{ 
    \left\langle 
    \frac{\boldsymbol{T'}}{\|\boldsymbol{T'}\|} \cdot \frac{\Phi(\boldsymbol{s}_i^{j})}{\|\Phi(\boldsymbol{s}_i^{j})\|} 
    \right\rangle \;\middle|\; i = 1, \dots, m;\; j = 1, \dots, k_{s} 
    \right\}.
\end{equation}
By indexing $S^{s}_{\text{tool}}$, we obtain the sections $\mathcal{S}_{\text{tool}}$ returned by the corresponding tool and concatenate them to form the final search result $\mathcal{S}_{\text{search}}$. Subsequently, $\mathcal{S}_{\text{search}}$ is fed into VLM-PRF along with the top-$k$ article $D$, which is retrieved by directly searching the input image. Following the specification in Figure~\ref{fig:tool}, $\mathcal{S}_{\text{search}}$ is filled in $<$search\_result$>$, $D$ is filled in $<$retrieved\_information$>$, where //$...$// represents long text.
\subsection{Filtering Stage}
During the retrieval processing, a large amount of redundant information is generated, with only a small fraction containing key details relevant to answering the question.
Previous approaches~\cite{yan2024echosight,cocchi2024augmenting} attempt to mitigate this by reranking and selecting more relevant passages. However, article- or section-level reranking methods can only filter at the passage level, often retaining significant noise.
To address this limitation, we propose training the model using reinforcement learning, guided by answer accuracy. 
This approach enables the model to filter retrieval results in a question-specific manner, reducing the influence of irrelevant content. Specifically, Wiki-PRF guides VLM-PRF to process the directly retrieved information $D$, comprising both image-derived data from $I$ and search results $\mathcal{S}_{\text{search}}$, and output its reasoning within \textcolor{teal}{<think>} and \textcolor{teal}{</think>} tags. The model then generates a compact, task-oriented knowledge representation $F$ within \textcolor[rgb]{0.8,0,0}{<answer>} and \textcolor[rgb]{0.8,0,0}{</answer>} tags. 
\begin{equation}
    F = \text{VLM}\text{-}\text{PRF}(D,\mathcal{S}_{\text{search}}), 
\end{equation}
\begin{equation}
    \quad A = \text{VLM}(F,Q),
\end{equation}
where $D$ denotes the retrieved information corresponding to $I$, $\mathcal{S}_{\text{search}}$ represents the external search results obtained via tool-based retrieval, and $F$ is the filtered, task-oriented knowledge representation produced by the reinforcement learning module. After generating task-oriented knowledge $F$, Wiki-PRF uses context to generate the final answer A via the VLM.

\subsection{Training via Reinforcement Learning}

To improve the model's tool selection and information filtering strategies, we utilize GRPO~\cite{shao2024deepseekmath} with removed KL divergence constraint for VLM-PRF training.
The formula can be defined as:
\begin{equation}
\resizebox{1.0\linewidth}{!}{$
\displaystyle 
\mathcal{J}(\theta) = 
\mathbb{E}_{q \sim P(Q), \{o_i\}_{i=1}^G \sim \pi_{\theta_{\text{old}}}(O|q)} 
\left[ 
\frac{1}{G} \sum_{i=1}^G \frac{1}{|o_i|} \sum_{t=1}^{|o_i|} 
\min \left( 
r_{i,t}(\theta) \hat{A}_{i,t},\ 
\text{clip}(r_{i,t}(\theta), 1-\epsilon, 1+\epsilon) \hat{A}_{i,t}
\right)
\right],
$}
\end{equation}
\begin{equation}
r_{i,t}(\theta) = \dfrac{\pi_\theta(o_{i,t}|q, o_{i,<t}; R(q))}{\pi_{\theta_{\text{old}}}(o_{i,t}|q, o_{i,<t}; R(q))},
\end{equation}
where  $\{o_i\}_{i=1}^G$ denotes the $G$ responses generated for question $Q$, and $|o_i|$ is the length of the $i$-th response.  
The term $\pi_\theta(o_{i,t}|q, o_{i,<t}; R(q))$ represents the conditional probability of token $o_{i,t}$ at decoding step $t$, conditioned on previous tokens $o_{i,<t}$ and retrieved information $R(q)$.

Besides, we design a reward function as the primary supervisory signal to guide the model in enhancing its tool invocation and its filtering of retrieved information. Specifically, we employ a format-based reward to encourage VLM-PRF to perform multi-step reasoning about tool usage within the \textcolor{teal}{<think>} tags, make appropriate tool calls within the \textcolor{blue}{<tool>} tags, and further process the retrieved information within the \textcolor{teal}{<think>} tags. Finally, the refined and filtered results are output within the \textcolor[rgb]{0.8,0,0}{<answer>} tags.
Furthermore, we introduce an answer reward to supervise the content generated within the \textcolor[rgb]{0.8,0,0}{<answer>} tags, ensuring that the model produces high-quality, relevant, and well-structured responses. The final reward function can be presented by:
\begin{equation}
    r_{\phi}(x,y) = \alpha EM(a_{\text{pred}},a_{\text{gt}})+\beta M(a_{\text{tool}},t_{\text{tool}})+\gamma M(a_{\text{filter}},t_{\text{filter}}),
\end{equation}
\begin{equation}
    M(x,y) = \begin{cases}
    1 & \text{if }  \text{match}, \\
    0      & \text{if } \text{unmatch},
\end{cases}
\end{equation}
where $r_{\phi}(x,y)$ represents the reward between input $x$ and output $y$; $EM$ denotes the evaluation function for answers, such as exact matching; $M$ employs regular expression matching to verify format compliance. $\alpha$, $\beta$, and $\gamma$ are the weighting coefficients for these components, with values of 1, 0.3, and 0.7 respectively in our method. Moreover, $a_{\text{pred}}$ refers to the model's output answer, $a_{\text{gt}}$ represents the ground truth, while $a_{\text{tool}}$ and $a_{\text{filter}}$ correspond to the model's outputs during the processing and filtering stages. $t_{\text{tool}}$ and $t_{\text{filter}}$ represent the templates for tool usage and filtering.

\section{Experiments}
\subsection{Experimental Setup}
\textbf{Datasets.} 
We evaluated our experimental results on two main datasets: InfoSeek~\cite{lin2022retrieval} and Encyclopedic VQA (E-VQA)\cite{lin2023fine}. 
(1) InfoSeek\cite{lin2022retrieval} contains 1.3M VQA pairs matched to 11K images from OVEN\cite{hu2023open}. The training set (934K) and validation set (73K) are strictly divided by both entities and questions. The validation set is further categorized into two types: Unseen Entity and Unseen Question. Following the setup of ~\cite{yan2024echosight}, we used a knowledge base consisting of 100K Wikipedia entries and reported evaluation results on the entire validation split.
(2) E-VQA~\cite{lin2023fine} consists of over 221K unique question-answer pairs, each associated with up to five images sourced from iNaturalist~\cite{van2021benchmarking} and the Google Landmarks Dataset v2~\cite{weyand2020google}. The dataset includes two types of questions: Single-hop and Two-hop. The samples are divided into training, validation, and test sets with 1M, 13.6K, and 5.8K items respectively. Like other methods~\cite{yan2024echosight,cocchi2024augmenting}, we report our results on the 5.8K test set. We also evaluate OK-VQA~\citep{marino2019ok}, a 14K-question dataset spanning diverseknowledge domains.

\textbf{Baselines. } 
To validate the effectiveness of our Wiki-PRF, we establish two baselines: (1) Base Model: Directly answer questions without any RAG pipelines. (2) Wiki-PRF without RL: Answer questions with our Wiki-PRF method before RL fine-tuning. 
These baselines serve to assess the contributions of our three-stage Wiki-PRF design and the benefits of RL fine-tuning, respectively.

\textbf{Evaluation Metrics. } 
In the KB-VQA task, we focus on evaluating both retrieval and QA metrics. For the retrieval metric, we use recall to determine whether the correct article appears among the top-k retrieved results. For the QA metric, following the original dataset settings, we apply VQA accuracy\cite{goyal2017making,marino2019ok} for InfoSeek and BEM score\cite{zhangbertscore} for E-VQA.

\textbf{Implementation Details.} 
Given the strong vision-language understanding capabilities of the Qwen2.5-VL series, we adopt Qwen2.5-VL-3B and Qwen2.5-VL-7B as our base models. We apply GRPO for reinforcement learning. Specifically, we set the number of generations to 8, the sampling temperature to 0.7, the number of training epochs to 2, and the learning rate to 1$e$-5. We utilize LoRA-based fine-tuning, with the LoRA rank to 64, the LoRA alpha to 128, and the dropout rate to 0.05.
For the retriever, we use a frozen EVA-CLIP 8B model for both retrieval and similarity computation. Image features are indexed and retrieved using cosine similarity with the Faiss-GPU library. 
By default, we use the Top-1 article retrieved from image retrieval alongside the Top-5 articles identified through tool calls. All sections from the image-retrieved article are retained, while the top-k sections from each of the tool-retrieved articles are kept. This combined information is then provided to the next stage for filtering.
The total training process takes approximately 15 hours using 8 A800 GPUs. 
Our framework, Wiki-PRF, is implemented in two configurations: \textbf{Wiki-PRF-3B} and \textbf{Wiki-PRF-7B}. Models further fine-tuned using reinforcement learning are referred to as \textbf{VLM-PRF-3B} and \textbf{VLM-PRF-7B}.

\begin{table*}[t]
  \centering
  \setlength{\tabcolsep}{.4em}
  \resizebox{0.97\linewidth}{!}{
  \begin{tabular}{lc c c c cc c ccc}
   \toprule
    & & & & & \multicolumn{2}{c}{\textbf{E-VQA}} & & \multicolumn{3}{c}{\textbf{InfoSeek}} \\
    \cmidrule{6-7} \cmidrule{9-11}
     \textbf{Method} & \textbf{Model} & & \textbf{Retriever} & & Single-Hop & All & & Unseen-Q & Unseen-E & All \\
    \midrule
    \rowcolor{lightgray} 
    \multicolumn{11}{l}{\textit{Zero-shot MLLMs}} \\
    BLIP-2~\cite{li2023blip} & Flan-T5$_\text{XL}$ & & - & &  12.6 & 12.4 & & 12.7 & 12.3 & 12.5 \\
    InstructBLIP~\cite{dai2023instructblip} & Flan-T5$_\text{XL}$ & & - & &  11.9 & 12.0 & & 8.9 & 7.4 & 8.1 \\
    LLaVA-v1.5~\cite{liu2024improved} & Vicuna-7B & & - & & 16.3 & 16.9 & & 9.6 & 9.4 & 9.5 \\
    GPT-4V~\cite{achiam2023gpt} & - & & - & & 26.9 & 28.1 & & 15.0 & 14.3 & 14.6 \\
    \rowcolor{cyan!15}
    Qwen2.5-VL-3B (Base)~\cite{bai2025qwen2} & - & - & - & & 17.9 & 19.6 & & 20.4 & 21.9 & 21.4 \\
    \rowcolor{cyan!15}
    Qwen2.5-VL-7B (Base)~\cite{bai2025qwen2} & - & - & - & & 21.7 & 20.3 & & 22.8 & 24.1 & 23.7 \\
    \midrule
    \rowcolor{lightgray} 
    \multicolumn{11}{l}{\textit{Retrieval-Augmented Models}} \\
    DPR$_\text{V+T}$~\cite{lerner2024cross}$^\dagger$ & Multi-passage BERT & & CLIP ViT-B/32 & & 29.1 & - & & - & - & 12.4 \\
    RORA-VLM~\cite{Qi2024RoRAVLMRR}$^\dagger$ & Vicuna-7B & & CLIP+Google Search & & - & 20.3 & & 25.1 & 27.3 & - \\
    EchoSight~\cite{yan2024echosight}$^\dagger$ & Mistral-7B/LLaMA-3-8B & & EVA-CLIP-8B & & 19.4 & - & & - & - & 27.7 \\
    Wiki-LLaVA~\cite{caffagni2024wiki} & Vicuna-7B & & CLIP ViT-L/14+Contriever & & 17.7 & 20.3 & & 30.1 & 27.8 & 28.9 \\
    ReflectiVA~\cite{cocchi2024augmenting} & LLaMA-3.1-8B & & EVA-CLIP-8B & & 28.0 & 29.2 & & 40.4 & 39.8 & 40.1 \\
    MMKB-RAG~\cite{ling2025mmkb} & Qwen2-7B & & EVA-CLIP-8B & & \textbf{39.7} & 35.9 & & 36.4 & 36.3 & 36.4 \\
    \rowcolor{cyan!15}
    VLM-PRF (w/o RL) & Qwen-2.5VL-3B & & EVA-CLIP-8B & & 26.6 &25.6 & & \text{34.2} & \text{33.7} & \text{34.0} \\
    \rowcolor{cyan!15}
    VLM-PRF (w/o RL) & Qwen-2.5VL-7B & & EVA-CLIP-8B & & 28.9 & 28.6 & & \text{40.0} & \text{39.4} & \text{39.5} \\
    \midrule
    \rowcolor{lightgray} 
    \multicolumn{11}{l}{\textit{Retrieval-Augmented Models with Reinforcement Learning}} \\
    \rowcolor{cyan!15}
    VLM-PRF (\textbf{Ours}) & LLaMA-3.1-8B & & EVA-CLIP-8B & & 36.3 &35.5 & & 41.3 & 40.6 & 40.8 \\
    \rowcolor{cyan!15}
    VLM-PRF (\textbf{Ours}) & Qwen-2.5VL-3B & & EVA-CLIP-8B & & 31.1 & 32.4 & & 39.7 & 38.8 & 39.0 \\
    \rowcolor{cyan!15}
    VLM-PRF (\textbf{Ours}) & Qwen-2.5VL-7B & & EVA-CLIP-8B & & 37.1 &\underline{36.0} & & \underline{43.3} & \textbf{42.7} & \textbf{42.8}\\
    \rowcolor{cyan!15}
    VLM-PRF (\textbf{Ours}) & InternVL3-8B & & EVA-CLIP-8B & & \textbf{40.1} &\textbf{39.2} & & \textbf{43.5} & \underline{42.1} & \underline{42.5}\\
  \bottomrule
  \end{tabular}
  }
\caption{\textbf{VQA accuracy on E-VQA and InfoSeek.} The metrics of baselines and our methods are highlighted in light blue. $\dagger$ indicates results that are not directly comparable due to different knowledge bases.
}
\label{tab:results}
\vspace{-0.35cm}
\end{table*}
\begin{figure}[htbp]
    \centering
    \includegraphics[width=1.0\linewidth]{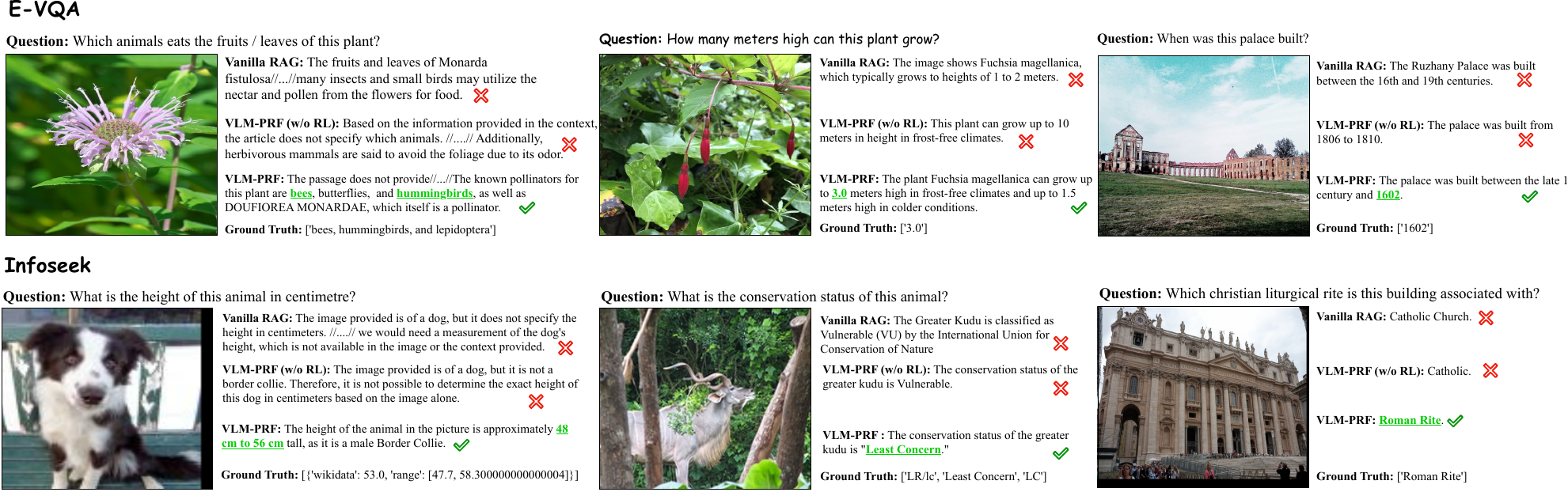}
    \caption{\textbf{Qualitaive examples of Wiki-PRF.}
    }
    \label{fig: case study}
    \vspace{-0.3cm}
\end{figure}
\subsection{Main Results}
\textbf{VQA Results.} We assess our method on the aforementioned VQA datasets, comparing it against various MLLMs and retrieval-augmented approaches as shown in Table~\ref{tab:results}. 
Among zero-shot MLLM approaches, we found that without RAG, the MLLM model achieves only relatively low accuracy in answering questions, highlighting the challenges presented by the KB-VQA task.
Regarding retrieval-augmented methods, our Wiki-PRF before RL training, achieves an accuracy of 34.0 with the 3B model and 39.5 with the 7B model in InfoSeek dataset, surpassing several well-trained methods such as EchoSight~\cite{yan2024echosight} and Wiki-LLaVA~\cite{caffagni2024wiki}. This finding further confirms the effectiveness of our proposed three-stage approach.
After RL training, our Wiki-PRF-7B establishes a new state-of-the-art accuracy of 36.0 on E-VQA and 42.8 on InfoSeek, outperforming all previous methods, including ReflectiVA~\cite{cocchi2024augmenting} and MMKB-RAG~\cite{ling2025mmkb}. Moreover, our method achieves consistent improvements across different base models, setting a new SOTA score of 39.2\% on E-VQA with InternVL3-8B as the base model. This underscores the impact of reinforcement learning in enhancing the model's RAG capabilities.

\begin{table}[!htbp]
\centering
\begin{minipage}{0.4\textwidth}
\centering
\caption{\textbf{Recall of retrieved articles.}
}
\label{tab:recall_results}
\resizebox{1.0\textwidth}{!}{
\begin{tabular}{lcc}
\toprule
Model & Retrieval Input & Retrieval Recall \\
\midrule
None & images & 45.56 \\
Qwen2.5-VL-3B & images + tools & 48.32 \\
Qwen2.5-VL-7B & images + tools  & 53.44 \\
VLM-PRF-3B & images + tools & 50.16 \\
VLM-PRF-7B & images + tools & 54.89 \\
\bottomrule
\end{tabular}}
\end{minipage}%
\hspace{0.05\textwidth}
\begin{minipage}{0.5\textwidth}
\centering
\caption{\textbf{Tool usage statistics.} Statistical analysis of tool usage (mean and variance).}
\label{tab:tool_calls}
\resizebox{\textwidth}{!}{
\begin{tabular}{@{}l*{4}{c}@{}}
\toprule
Model & Combinations & Captioning & Grounding & Flipping \\
\midrule
Qwen2.5-VL-3B & 34 & 0.86 / 1.10 & 0.40 / 0.50 & 0.04 / 0.22 \\
Qwen2.5-VL-7B & 34 & 2.43 / 1.03 & 0.85 / 0.28 & 0.22 / 0.42 \\
VLM-PRF-3B & 53 & 1.52 / 1.22 & 0.54 / 0.50 & 0.15 / 0.36 \\
VLM-PRF-7B & 40 & 2.43 / 1.13 & 0.93 / 0.36 & 0.26 / 0.44 \\
\bottomrule
\end{tabular}}
\end{minipage}
\end{table}

\textbf{Effectiveness of the Processing Stage.} 
Table~\ref{tab:recall_results} presents the recall of articles retrieved from InfoSeek under various settings. The Top-1 retrieval performance using direct image retrieval is 45.56\%. Combined with Top-5 article retrievals from our tools, this rate increases to 48.32\% and 53.44\%. Further improvement with reinforcement learning supervision raises it to 50.16\% and 54.89\%. Notably, we observe that models trained with RL supervision exhibit greater diversity and frequency in tool selection, as shown in Table~\ref{tab:tool_calls}. Specifically, the combinations of tool calls increase after RL training, demonstrating that the model can dynamically and flexibly construct tool invocation schemes. At the same time, the captioning tool is invoked most frequently, highlighting its role as the most common and direct tool for enhancing article recall. Overall, this demonstrates that RL can encourage the model to leverage a broader array of tools by optimizing for the final answer reward, thereby showcasing its flexibility.

\begin{table}[!htbp]
\centering
\begin{minipage}{0.4\textwidth}
\centering
\caption{\textbf{Performance on OK-VQA.}
}
\label{tab:OK-VQA dataset}
\resizebox{1.0\textwidth}{!}{
    \begin{tabular}{llll}
        \toprule
        Method &Model &OK-VQA \\
        \midrule
        Qwen2.5-VL-3B &- &62.1\\
        Qwen2.5-VL-7B &- & 72.4\\
        KU-RAG~\cite{zhang2025fine} &LLaVA-Next-7B &73.1\\
        MMKB-RAG~\cite{ling2025mmkb} &LLaMA-3.1-8B &65.4\\
        Wiki-PRF-3B &VLM-PRF-3B &\textbf{68.6}\\
        Wiki-PRF-7B &VLM-PRF-7B &\textbf{77.8}\\
         \bottomrule
    \end{tabular}}
\end{minipage}%
\hspace{0.05\textwidth}
\begin{minipage}{0.5\textwidth}
\centering
\caption{\textbf{Filtering from oracles.} VQA Accuracy in Oracle Setting with Ground-Truth Articles.}
\label{tab:oracle_results}
\resizebox{\textwidth}{!}{
\begin{tabular}{lcc}
\toprule
Method & Model & VQA Accuracy \\
\midrule
Wiki-LLaVA~\cite{caffagni2024wiki} & Vicuna-7B & 51.5 \\
ReflectiVA~\cite{cocchi2024augmenting} & LLaMA-3.1-8B & 57.6 \\
Wiki-PRF-3B (\textbf{Ours}) & VLM-PRF-3B & 64.4 \\
Wiki-PRF-7B (\textbf{Ours}) & VLM-PRF-7B & 65.8 \\
\bottomrule
\end{tabular}}
\end{minipage}
\vspace{-5pt}
\end{table}


\begin{table}[!htbp]
\centering
\caption{\textbf{Comparison of SFT and RL.} We sampled 2,000 instances from InfoSeek and present a comparison between the results of SFT and RL.}
\label{tab:RL}
\resizebox{0.75\textwidth}{!}{
\begin{tabular}{lc  cc cc}
\toprule
Model &Unseen Question (UQ)	 &Unseen Entity (UE) & ALL \\ \midrule
Qwen2.5-VL-7B & 39.1  &40.5  &40.2  \\
Wiki-PRF-7B (SFT) & 41.5  &41.9  &41.8\\
Wiki-PRF-7B (RL) & \textbf{46.6}  &\textbf{46.2} &\textbf{46.3} \\
\bottomrule
\end{tabular}
}
\vspace{-8pt}
\end{table}

\textbf{Effectiveness of the Filtering Stage.} 
To evaluate whether our filtering stage can effectively extract accurate information from given sections, we conduct experiments similarly to other methods under an oracle setting, where the ground-truth entity (i.e., the Wikipedia page associated with the query) is provided. Compared with other methods, our final VQA accuracy is much higher (65.8\%) as shown in Table~\ref{tab:oracle_results}, the fine-tuned model can more efficiently locate the necessary information when given the oracle retrieval information.

\textbf{Effectiveness of Reinforcement Learning.} 
To investigate its effectiveness, we evaluated the use of supervised fine-tuning (SFT) for the filtering stage. Specifically, we trained a dedicated filtering model using SFT, keeping all other configurations identical to reinforcement learning (RL). The results are presented in Table~\ref{tab:RL}. The RL model significantly outperforms the SFT model on the test set. The reason is that SFT tends to imitate superficial patterns, which limits its generalization capability. In contrast, RL enables the model to understand the underlying principles of information filtering, leading to a far more robust and generalizable performance.

\textbf{Results on more benchmarks.}
As shown in Table~\ref{tab:OK-VQA dataset}, we evaluate our model on the widely used OK-VQA benchmark. We can see that our Wiki-PRF-7B achieves a new state-of-the-art score of 77.8 on OK-VQA. The consistent performance improvement on multiple benchmarks confirms our method's strong generalization capability.

\subsection{Ablation Studies}

\textbf{Modules and Tools.}
Table~\ref{tab:ablations} presents the ablation study of our carefully designed stages and tools based on 10K samples from the InfoSeek validation set. In the module ablation, VLM-PRF-3B model improves the baseline by 2.54\% and 2.02\% when combining the processing stage and filtering stage, respectively, underscoring the effectiveness of these two modules. In the tool ablation, the caption and grounding tools increase performance by 1.94\% and 0.98\%, respectively. The combined use of all tools yields a peak performance of 39.48\%. This result demonstrates that Wiki-PRF significantly boost the final VQA accuracy by adeptly utilizing tools to provide more relevant information.
\begin{table}[htbp]
\vspace{-10pt}
\centering
\caption{\textbf{Ablation studies on the scale of knowledge base on InfoSeek.}}
\label{tab:knowledge base}
\resizebox{0.6\textwidth}{!}{
\begin{tabular}{lc  cc cc cc}
\toprule
Method &Model &10k &50k & 100k \\ \midrule
Vanilla-MRAG & Qwen2.5-VL-3B &49.7 &32.1 &21.4  \\
Wiki-PRF-3B& VLM-PRF-3B  &53.0 &43.7 &39.0\\
Vanilla-MRAG & Qwen2.5-VL-7B & 56.3 &39.6 &23.7\\
Wiki-PRF-7B & VLM-PRF-7B & 60.3 & 51.2 & 42.8 \\
\bottomrule
\end{tabular}
}
\end{table}

\textbf{Scale of Knowledge Base.} 
In Table~\ref{tab:knowledge base}, we verfied that the correct document for each evaluation question was included in knowledge bases of all scales, to assess the impact of knowledge base size. The result demonstrates that both our method and the baseline exhibit performance degradation as knowledge base size increases. This occurs because larger knowledge bases introduce additional noise, increasing retrieval difficulty, which is a universal challenge for current RAG methods. Critically, Wiki-PRF demonstrates a significantly slower rate of degradation for both 3B and 7B models.

\begin{table}[htbp]
\centering
\caption{\textbf{Modules and tools ablation.}
"Processing", "Retrieval" and "Filtering" denote the three distinct stages proposed in our approach.}
\label{tab:ablations}
\resizebox{0.9\textwidth}{!}{
\begin{tabular}{lc c c c c}
\toprule
Model & Processing & Retrieval & Filtering & Tools & VQA Accuracy \\
\midrule
\rowcolor{lightgray}
\multicolumn{6}{l}{\textit{Modules Ablation}} \\
Qwen2.5-VL-3B &  & \ding{52} &  & - & 34.22 \\
VLM-PRF-3B &  & \ding{52} & \ding{52} & - & 36.24 \\
VLM-PRF-3B & \ding{52} & \ding{52} &  & Multi-Tools & 36.76 \\
VLM-PRF-3B & \ding{52} & \ding{52} & \ding{52} & Multi-Tools & 39.48 \\
\midrule
\rowcolor{lightgray}
\multicolumn{6}{l}{\textit{Tools Ablation}} \\
VLM-PRF-3B & \ding{52} & \ding{52} & \ding{52} & Grounding & 37.22 \\
VLM-PRF-3B & \ding{52} & \ding{52} & \ding{52} & Captioning & 38.18 \\
VLM-PRF-3B & \ding{52} & \ding{52} & \ding{52} & Multi-Tools & 39.48 \\
\bottomrule
\end{tabular}
}
\vspace{-5pt}
\end{table}

\begin{table*}[htbp]
\centering
\begin{tabular}{cc}
\begin{minipage}{0.45\linewidth}
\centering
\caption{\textbf{Ablation on training sample size.}}
\label{training_numbers}
\resizebox{\linewidth}{!}{
\begin{tabular}{l cc cc cc cc}
\toprule
VLM Model  & 2K & 4K & 6K & 8K \\
\midrule
VLM-PRF-3B & 37.30\% & 39.48\% & 38.92\% & 40.83\% \\
VLM-PRF-7B & 42.13\% & 43.10\% & 43.09\% & 43.80\% \\
\bottomrule
\end{tabular}
}

\end{minipage} &
\begin{minipage}{0.5\linewidth}
\centering
\caption{\textbf{Ablation on retrieved article quantity.}}
\label{tab:article_numbers}
\resizebox{\linewidth}{!}{
\begin{tabular}{ccccc}
\toprule
Retrieved Article & Top-1 & Top-3 & Top-5 & Top-7\\
\midrule
InfoSeek & 38.85\% & 39.10\% & 39.48\% & 39.57\% \\
\bottomrule
\end{tabular}
}
\end{minipage}
\end{tabular}
\end{table*}

\textbf{Training Samples.}
Table~\ref{training_numbers} presents the accuracy achieved with different training sample sizes. As the number of training samples increases, we observe a general upward trend in accuracy. Balancing accuracy with training time, we opt for 4K samples as our default experimental setting. Unlike other methods that necessitate training on the complete training dataset (InfoSeek: 934K, E-VQA: 1M), our approach attains comparable or even superior results using only a small subset of samples. 
We attribute this efficiency to reinforcement learning's ability to effectively stimulate the model to leverage tools for retrieving additional information and integrating existing knowledge to determine the correct answers, rather than relying on memorizing specific question-answer patterns.


\textbf{Retrieved Articles.}
Table~\ref{tab:article_numbers} illustrates the ablation of retrieving top-k  articles when utilizing tools. We evaluated the VQA accuracy of Wiki-PRF-3B on the InfoSeek dataset with varying numbers of retrieved articles: 1, 3, 5, and 7. The results indicate that as the number of retrieved articles increases, VQA accuracy tends to improve gradually, albeit with diminishing returns. To balance inference time and accuracy, we set the K=5 as the optimal choice and utilize the Top-3 relevant sections from these K=5 articles to supplement knowledge. More experiments and details about the retrieval of tool calls are presented in the supplementary materials.
\begin{table}[htbp]
\centering
\caption{\textbf{Inference time of each stage.} We sampled 1,000 instances from InfoSeek and measured the average duration of each stage per sample.}
\label{tab:Inference Time}
\resizebox{0.7\textwidth}{!}{
\begin{tabular}{lc  cc cc cc}
\toprule
Model &Processing \& Retrieval &Filtering &Answering & Total \\ \midrule
VLM-PRF-3B & 2.2s  &3.4s  &0.59s &6.23s  \\
VLM-PRF-7B& 3.3s  &4.6s  &0.74s &8.77s\\
\bottomrule
\end{tabular}
}
\end{table}

\textbf{Inference Time.} We evaluated the VLM-PRF model's stage-wise time cost per sample in Table~\ref{tab:Inference Time}. The results show that the Processing \& Retrieval stages and Filtering stage consume more time than the Answering stage.  This duration primarily stems from tool invocation and long-text processing, both of which can be further optimized in future improvements.

\section{Conclusion}
In this paper, we propose Wiki-PRF, a three-stage Process-Retrieval-Filtering framework that represents first reinforcement learning method for multimodal retrieval-augmented generation. By guiding models to invoke tools for processing raw information during the processing stage and filtering retrieved knowledge during the filtering stage, the trained VLM-PRF model significantly enhances performance on Knowledge-Based Visual Question Answering tasks. Extensive experiments demonstrate state-of-the-art results on E-VQA and InfoSeek benchmarks. While limited to three retrieval tools in this study, future work may explore expanded tool integration to further advance capabilities.

\section{Acknowledgements}
This work was supported by the Strategic Priority Research Program of Chinese Academy of Sciences (Grant No. XDA0480200),
the National Natural Science Foundations of China (Grant No.62306310).
\bibliography{reference}
\newpage

\newpage
\section*{NeurIPS Paper Checklist}

\begin{enumerate}

\item {\bf Claims}
    \item[] Question: Do the main claims made in the abstract and introduction accurately reflect the paper's contributions and scope?
    \item[] Answer: \answerYes{} 
    \item[] Justification: We present the motivation of our method as clearly as possible in the abstract and introduction.

\item {\bf Limitations}
    \item[] Question: Does the paper discuss the limitations of the work performed by the authors?
    \item[] Answer: \answerYes{} 
    \item[] Justification: We present the limitations of our approach at the end of the paper.

\item {\bf Theory assumptions and proofs}
    \item[] Question: For each theoretical result, does the paper provide the full set of assumptions and a complete (and correct) proof?
    \item[] Answer: \answerNA{} 
    \item[] Justification: The paper does not include theoretical results.

    \item {\bf Experimental result reproducibility}
    \item[] Question: Does the paper fully disclose all the information needed to reproduce the main experimental results of the paper to the extent that it affects the main claims and/or conclusions of the paper (regardless of whether the code and data are provided or not)?
    \item[] Answer: \answerYes{} 
    \item[] Justification: We present our method in sufficient detail in the Methods to allow readers to reproduce the experiments.

\item {\bf Open access to data and code}
    \item[] Question: Does the paper provide open access to the data and code, with sufficient instructions to faithfully reproduce the main experimental results, as described in supplemental material?
    \item[] Answer:  \answerYes{} 
    \item[] Justification: E-VQA and Infoseek are two public datasets we use. We will open source our code when appropriate.

\item {\bf Experimental setting/details}
    \item[] Question: Does the paper specify all the training and test details (e.g., data splits, hyperparameters, how they were chosen, type of optimizer, etc.) necessary to understand the results?
    \item[] Answer: \answerYes{} 
    \item[] Justification: We explain our training and testing numbers in detail in the Experimental Details.

\item {\bf Experiment statistical significance}
    \item[] Question: Does the paper report error bars suitably and correctly defined or other appropriate information about the statistical significance of the experiments?
    \item[] Answer: \answerYes{} 
    \item[] Justification: We describe the metrics we use and their significance in section metrics.

\item {\bf Experiments compute resources}
    \item[] Question: For each experiment, does the paper provide sufficient information on the computer resources (type of compute workers, memory, time of execution) needed to reproduce the experiments?
    \item[] Answer: \answerYes{} 
    \item[] Justification: We describe the devices we used and the computation time in the implementation details and further provide the inference time in the supplementary material.
    
\item {\bf Code of ethics}
    \item[] Question: Does the research conducted in the paper conform, in every respect, with the NeurIPS Code of Ethics \url{https://neurips.cc/public/EthicsGuidelines}?
    \item[] Answer: \answerYes{} 
    \item[] Justification: Yes, our paper complies with NeurIPS ethical requirements.

\item {\bf Broader impacts}
    \item[] Question: Does the paper discuss both potential positive societal impacts and negative societal impacts of the work performed?
        \item[] Answer: \answerYes{} 
    \item[] Justification: We explain our motivation and the benefits to the model in the abstract, which is consistent with the Broader impacts of LLM research.
    
\item {\bf Safeguards}
    \item[] Question: Does the paper describe safeguards that have been put in place for responsible release of data or models that have a high risk for misuse (e.g., pretrained language models, image generators, or scraped datasets)?
    \item[] Answer: \answerNA{} 
    \item[] Justification: We use public data sets and do not involve risk.

\item {\bf Licenses for existing assets}
    \item[] Question: Are the creators or original owners of assets (e.g., code, data, models), used in the paper, properly credited and are the license and terms of use explicitly mentioned and properly respected?
    \item[] Answer: \answerYes{} 
    \item[] Justification: We mainly base it on VLM-R1, whose code is licensed under CC-BY 4.0. https://github.com/om-ai-lab/VLM-R1.

\item {\bf New assets}
    \item[] Question: Are new assets introduced in the paper well documented and is the documentation provided alongside the assets?
    \item[] Answer: \answerYes{} 
    \item[] Justification: We use open source code on github and manage our code using github's open source license. 

\item {\bf Crowdsourcing and research with human subjects}
    \item[] Question: For crowdsourcing experiments and research with human subjects, does the paper include the full text of instructions given to participants and screenshots, if applicable, as well as details about compensation (if any)? 
    \item[] Answer: \answerNA{} 
    \item[] Justification: The paper does not involve crowdsourcing nor research with human subjects.

\item {\bf Institutional review board (IRB) approvals or equivalent for research with human subjects}
    \item[] Question: Does the paper describe potential risks incurred by study participants, whether such risks were disclosed to the subjects, and whether Institutional Review Board (IRB) approvals (or an equivalent approval/review based on the requirements of your country or institution) were obtained?
    \item[] Answer: \answerNA{} 
    \item[] Justification: Studies of retrieval enhancement generation do not involve subjects.

\item {\bf Declaration of LLM usage}
    \item[] Question: Does the paper describe the usage of LLMs if it is an important, original, or non-standard component of the core methods in this research? Note that if the LLM is used only for writing, editing, or formatting purposes and does not impact the core methodology, scientific rigorousness, or originality of the research, declaration is not required.
    \item[] Answer: \answerYes{} 
    \item[] Justification: We finetune Qwen2.5-VL 3B/7B in our method.

\end{enumerate}
\appendix
 This appendix presents additional materials and results. First, we describe the complete workflow of our method in Sec.~\ref{sec:A} to enhance comprehension. Then, we give further descriptions of our prompts in experiments in Sec.~\ref{prompt}. Next, we provide more ablation studies for Wiki-PRF in Sec.~\ref{Additional experiment}. Finally, a series of visual results are presented in Sec.~\ref{Qualitative Results}, and the broader impacts are discussed in Sec.~\ref{Broader Impacts}.
\section{Workflow of Wiki-PRF}
\label{sec:A}
We detail the complete workflow of Wiki-PRF below.
\begin{itemize}
    \item \textbf{Processing Stage} 
    \begin{itemize}
        \item \textbf{Query Anysis:} Given the reference image $I$ and question $Q$, Wiki-PRF begins by analyzing the key information needed to solve the problem in \textcolor{teal}{<think>} and \textcolor{teal}{</think>} and subsequently specifies the required tools using the \textcolor{blue}{<tool> Tool: Content </tool>} format.
        \item \textbf{Tool Calling:} Upon capturing a tool request, Wiki-PRF parses tools enclosed in \textcolor{blue}{<tool>} and \textcolor{blue}{</tool>} tags and sequentially executes the corresponding functions.
        \begin{itemize}
            \item For Captioning, Wiki-PRF feeds the content following the caption to VLM-PRF to generate the retrieval query $Query_{\text{captioning}}$.
        \end{itemize} 
        \begin{itemize}
            \item For Grounding, Wiki-PRF first obtains the object coordinates from VLM-PRF, followed by performing the image cropping operation based on the coordinates. The resulting cropped image is then returned as the retrieval query $Query_{\text{grounding}}$.
        \end{itemize} 
        \begin{itemize}
            \item For Flipping, VLM-PRF directly returns the flipped image $I_{\text{flip}}$.
        \end{itemize}
        
    \end{itemize}
    \item \textbf{Retrieval Stage:} In the retrieval stage, Wiki-PRF follows a two-step process: it first retrieves the top-k articles $D$ based on the reference image $\mathcal{I}$, and then conducts further searches using the queries returned by the tools.
    \begin{itemize}
        \item \textbf{Captioning Search:} Given $Query_{\text{captioning}}$, Wiki-PRF initially retrieves the top $k$ most similar images and their associated documents from the knowledge base. These documents are then segmented into sections denoted as $\mathcal{S}_{\text{captioning}}$. Subsequently, Wiki-PRF computes the similarity between $Query_{\text{captioning}}$ and each section in $\mathcal{S}_{\text{captioning}}$, and selects the top-$k_{s}$ most relevant sections as the final retrieval results.
        \item \textbf{Grounding Search:} Given $Query_{\text{grounding}}$, same as Captioning Search, Wiki-PRF follows a procedure similar to that of captioning search by first retrieving the sections $\mathcal{S}_{\text{grounding}}$. The key difference lies in the subsequent step, where the Wiki-PRF computes the similarity between the question $Q$ and each section in $\mathcal{S}_{\text{grounding}}$. Finally, top-$k_{s}$ sections are selected as the retrieval results.
        \item \textbf{Constructing Search Result:} Wiki-PRF takes the union of all retrieval results, and then concatenates the sections in the union as $\mathcal{S}_{\text{search}}$.
    \end{itemize}
    \item \textbf{Filtering Stage:} Given the documents $D$ and the sections $\mathcal{S}_{\text{search}}$, Wiki-PRF leverages VLM-PRF to filter relevant information guided by the reference image $I$ and question $Q$. The reasoning process of VLM-PRF is presented within \textcolor{teal}{<think>} and \textcolor{teal}{</think>}, while the resulting task-oriented knowledge $F$ is output within \textcolor[rgb]{0.8,0,0}{<answer>} and \textcolor[rgb]{0.8,0,0}{</answer>}.
    \item \textbf{Answering:} With the task-oriented knowledge $F$, Wiki-PRF generates the final answer $A$.
\end{itemize} 
\section{Prompts Details in Wiki-PRF}
\label{prompt}
\subsection{Processing Stage}
\textbf{Prompt for Tool Calling:}
\userprompt{
Given a question whose answer is within a knowledge base, you need to utilize one or more following tools to query the knowledge base by providing information you need: $\backslash$'caption$\backslash$': Provide a detailed description related to the question, and the information will be used to query the external knowledge base to retrieve relevant knowledge points. $\backslash$'grounding$\backslash$': Identify the specific core subject related to the question and it will return concrete details about the area. $\backslash$'Flip$\backslash$': Flip the image left or right. Enclose your reasoning process within <think> and </think> without detailed illustrations, and specify the tools and contents you use within <tool> and </tool> to aid in querying the external knowledge base. Example: <think>reasoning process</think> <tool> \\
1. Flip: Flip left. \\
2. grounding: The panda on the tree. \\
3. caption: A panda is climbing the tree with a bird beside it.\\</tool> Here is the user question, \{Question\}.
}
\textbf{Prompt for Captioning:}
\userprompt{
Here is the question, \{Question\}. Here is the caption, \{Caption\}. describe the image in the context of the question and the caption."
}

\textbf{Prompt for Grounding:}
\userprompt{
"Locate \{object\}, output its bbox coordinates using JSON format."
}
\subsection{Filtering Stage}
\userprompt{
"Here is the user question, <question> \{Question\} </question>. Here is the relevant information retrieved through image retrieval, <retrieved\_information> \{Document\} </retrieved\_information>. Here is the relevant information through <tool>\{Search\}</tool>, <search\_result>\{Search\_result\}</search\_result>. \\To obtain useful information, you must conduct reasoning inside <think> </think> first every time you get new retrieved information. After reasoning, you should provide the filtered information inside <answer> and </answer>, without detailed illustrations."
}
\subsection{Prompt for Answer}
\userprompt{
"Here is the question, \{Question\}. Here is the retrieval information,\{Search$\_$results\}, short answer:"
}
\section{Additional Experiments}
\label{Additional experiment}
\subsection{Training Loss}
\begin{figure}[h]
    \centering
    \includegraphics[width=1\linewidth]{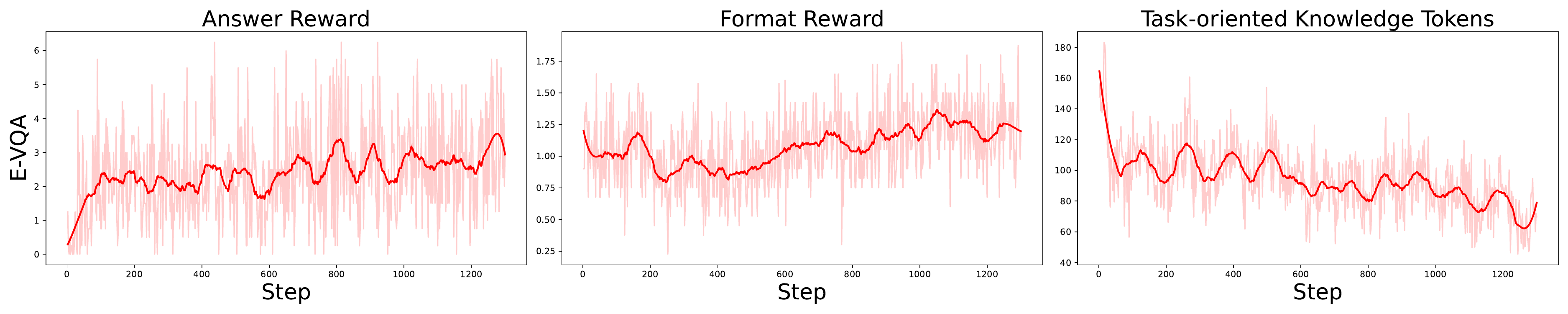}
    \caption{\textbf{The training curve of VLM-PRF-7B in E-VQA.}}
    \label{reinforcement learning loss}
\end{figure}

In this section, we present the training curve of VLM-PRF-7B under reinforcement learning in E-VQA. Figure~\ref{reinforcement learning loss} displays three key metrics: answer reward, format reward, and task-oriented knowledge tokens. As shown in Figure~\ref{reinforcement learning loss}, both the answer reward and format reward exhibit a consistent upward trend, indicating that as the model learns to invoke tools and filter relevant information, its accuracy in answering knowledge-based VQA questions gradually improves. This clearly demonstrates the effectiveness of GRPO in enhancing the model's RAG capabilities. 

Moreover, the tokens of the task-oriented knowledge decreases progressively with the number of training steps. This phenomenon suggests that the model becomes increasingly adept at identifying and retaining only the most relevant knowledge during the learning process. 

\subsection{Tool Inference Time}
 In Table~\ref{tab:Tool Time}, we analyze the execution time of individual tools under varying numbers of recalled articles (i.e., 3, 5, and 7). The results reveal that grounding requires more time than captioning. This can be attributed to the image processing operations involved in grounding, which result in increased computational demands during execution.
\begin{table}[h]
\centering
\caption{\textbf{Tool Calling Time Per Sample.}}
\label{tab:Tool Time}
\resizebox{0.6\textwidth}{!}{
\begin{tabular}{lc  cc cc cc}
\toprule
Model &Recall Numbers &Captioning &Grounding \\ \midrule
 & 3  &0.99s  &1.64s  \\
VLM-PRF-3B& 5  &1.07s  &1.69s\\
& 7  &1.11s  &1.73s\\
\bottomrule
\end{tabular}
}
\end{table}

\subsection{Weights of Rewards}
As shown in Table~\ref{tab:format:accuracy}, we investigate the influence of varying the weight of answer reward (i.e., $\alpha$) and the weight of format reward (i.e., $\beta+\gamma$) in the overall objective function on the InfoSeek dataset. We fix the values of $\beta$ and $\gamma$ to 1.0, and continuously adjust the ratio of $\alpha$ and $(\beta+\gamma)$.  By gradually decreasing $\alpha : (\beta+\gamma)$ from $3:1$ to $1:3$, we observe that the optimal performance is achieved when both components are equally weighted. Consequently, in our experiments, we adopt an equal ratio, where $\alpha=2.0$, $\beta=1.0$, and $\gamma=1.0$. 
\begin{table}[h]
    \centering
    \caption{\textbf{Ratios of Answer Reward and Format Reward Weights.}}
    \begin{tabular}{ccccccc}
        \toprule
        Model &3:1 & 2:1 & 1:1 & 1:2 &1:3 \\
        \midrule
        VLM-PRF-3B &38.80 &38.20  &39.48 &38.89 &38.53\\
         \bottomrule
    \end{tabular}
    \label{tab:format:accuracy}
\end{table}
\subsection{The Number of Selected Sections}
In Table~\ref{tab:Infoseek articles} and Table~\ref{tab:E-VQA articles}, we present ablation studies conducted on VLM-PRF-3B to evaluate the impact of varying the number of retrieved articles and sections during tool-based retrieval on the InfoSeek and E-VQA datasets. The tables report the final accuracy of Wiki-PRF-3B when top-1 and top-3 retrieved articles or sections are used during training.

The results indicate that the model performance generally improves as the number of selected sections increases. However, when only a single article is considered, the overall relevance of the article becomes the primary determinant of accuracy. The inclusion of redundant sections introduces noise and may lead to a decline in performance.

\begin{table}[h]
\centering
\begin{minipage}{0.45\textwidth}
\centering
\caption{\textbf{Retrieved Settings Ablation on Infoseek.}
}
\label{tab:Infoseek articles}
\resizebox{1.0\textwidth}{!}{
\begin{tabular}{cccc}
     \toprule
     Retrieved Settings & Top-1 Section & Top-3 Sections \\
     \midrule
     Top-1 Article & 38.96\% & 38.85\%  \\
     Top-3 Articles & 39.03\% & 39.10\%  \\
     Top-5 Articles & 39.39\% & 39.48\% \\
     \bottomrule
     \end{tabular}}
\end{minipage}%
\hspace{0.05\textwidth}
\begin{minipage}{0.45\textwidth}
\centering
\caption{\textbf{Retrieved Settings Ablation on E-VQA.}}
\label{tab:E-VQA articles}
\resizebox{\textwidth}{!}{
\begin{tabular}{cccc}
     \toprule
     Retrieved Settings & Top-1 Section & Top-3 Sections \\
     \midrule
     Top-1 Article & 24.28\% & 24.31\%  \\
     Top-3 Articles & 28.15\% & 28.94\%  \\
     Top-5 Articles & 32.10\% & 32.38\% \\
     \bottomrule
     \end{tabular}}
\end{minipage}
\end{table}


\section{Qualitative Results}
\label{Qualitative Results}
\subsection{Comparison of Wiki-PRF}
\begin{figure}[b]
    \centering
    \includegraphics[width = 1\linewidth]{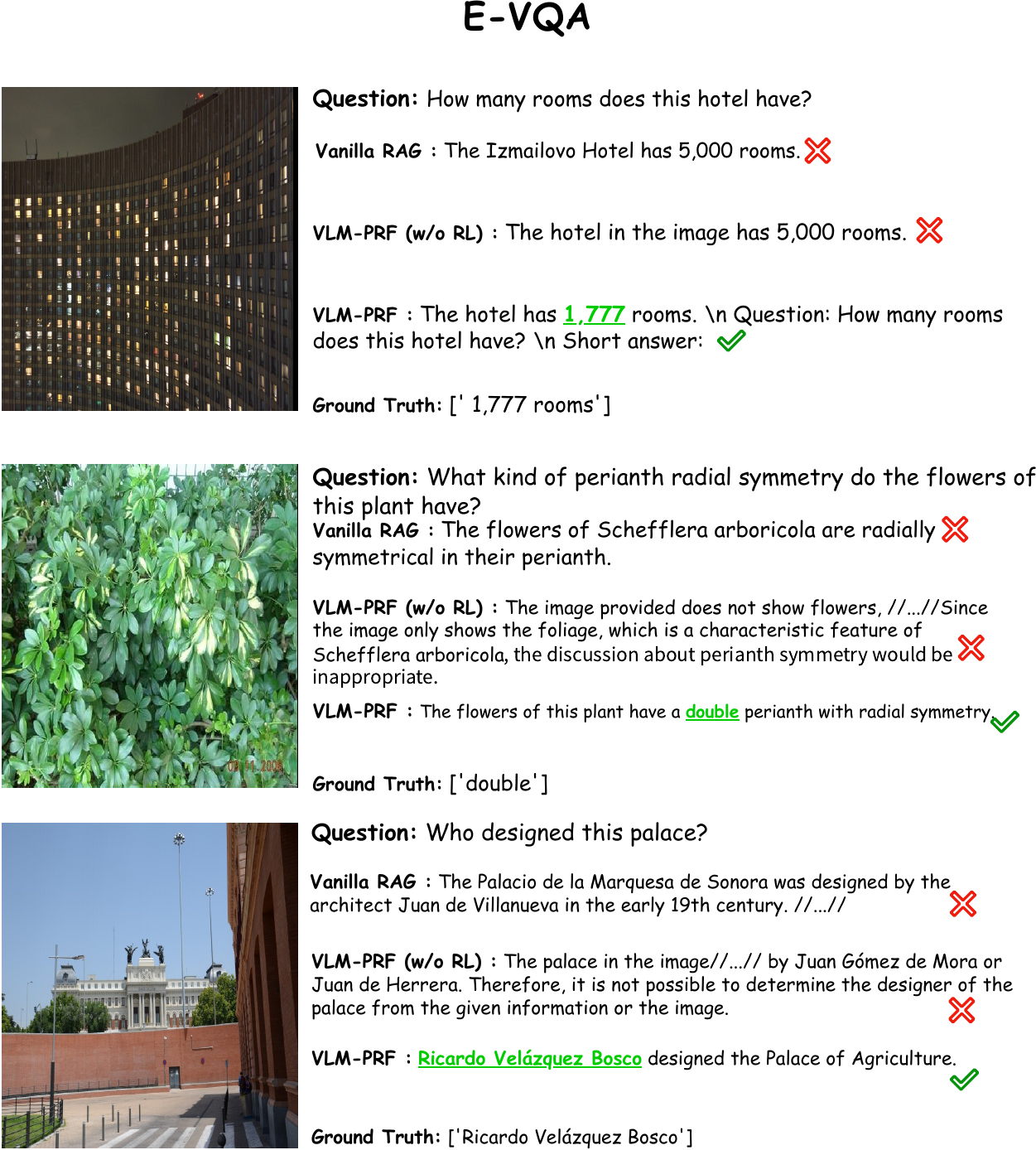}
    \caption{\textbf{Comparison on E-VQA}}
    \label{fig:case1}
\end{figure}

\begin{figure}
    \centering
    \includegraphics[width = 1\linewidth]{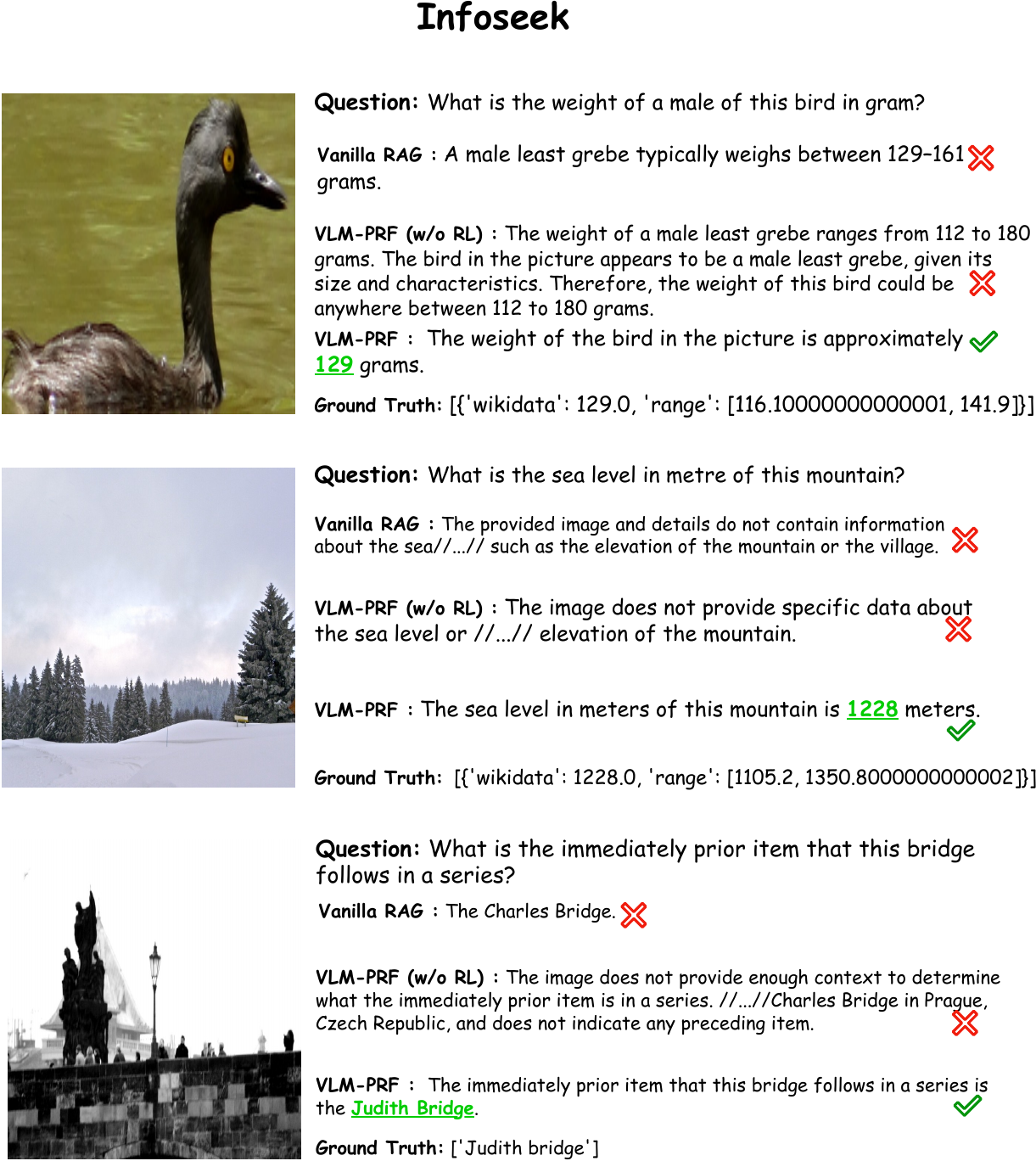}
    \caption{\textbf{Comparison on InfoSeek}}
    \label{fig:case2}
\end{figure}

We conduct a comparison between our method and two baselines: Vanilla RAG and Wiki-PRF without the reinforcement learning fine-tuning (Wiki-PRF w/o RL).
As shown in Figure~\ref{fig:case1}, we present a comparison across various scenes, including plants, buildings, and animals. Examples 2 and 3 in Figure~\ref{fig:case1} and example 1 in Figure~\ref{fig:case2} demonstrate the accuracy of our method in answering number-related questions. Examples 3 and 4 in Figure~\ref{fig:case1} show that our method can still accurately answer questions when the target subject is far away. The comparison results fully illustrate the effectiveness of our method.

\subsection{Illustration of Wiki-PRF on Variours Questions}
\begin{figure}
    \centering
    \includegraphics[width=1.1\linewidth]{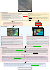}
    \caption{\textbf{Illustration of Wiki-PRF on Question E-VQA\_114 from E-VQA.}}
    \label{fig:case 1}
\end{figure}

\begin{figure}
    \centering
    \includegraphics[width=1.1\linewidth]{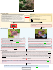}
    \caption{\textbf{Illustration of Wiki-PRF on Question E-VQA\_1182 from E-VQA.}}
    \label{fig:case 2}
\end{figure}

\begin{figure}
    \centering
    \includegraphics[width=1.1\linewidth]{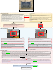}
    \caption{\textbf{Illustration of Wiki-PRF on Question E-VQA\_1747 from E-VQA.}}
    \label{fig:case 3}
\end{figure}

\begin{figure}
    \centering
    \includegraphics[width=1.1\linewidth]{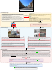}
    \caption{\textbf{Illustration of Wiki-PRF on Question Infoseek\_00012299 from Infoseek.}}
    \label{fig:case 4}
\end{figure}

\begin{figure}
    \centering
    \includegraphics[width=1.1\linewidth]{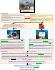}
    \caption{\textbf{Illustration of Wiki-PRF on Question Infoseek\_00033513 from Infoseek.}}
    \label{fig:case 5}
\end{figure}

\begin{figure}
    \centering
    \includegraphics[width=1.1\linewidth]{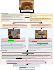}
    \caption{\textbf{Illustration of Wiki-PRF on Question Infoseek\_00005094 from Infoseek.}}
    \label{fig:case 6}
\end{figure}

In this section, we mainly show the case examples of Wiki-PRF. Figure~\ref{fig:case 1} and Figure~\ref{fig:case 4} show examples of visualizations of different tools working individually. Figure~\ref{fig:case 2} and Figure~\ref{fig:case 3} show scenarios where the two tools work together, but are called in different orders. Figure~\ref{fig:case 5} demonstrates that through tool calls, Wiki-PRF extends information retrieval to retrieve information that contributes to the answer. Figure~\ref{fig:case 6} shows an example of calling the captioning combination, proving that Wiki-PRF will make specific tool combinations according to the question. The filtered results across all samples are shorter than the original messages while retaining the correct answer, illustrating the effectiveness of the filtering stage.

\section{Broader Impacts of Wiki-PRF}
\label{Broader Impacts}
In this section, we focus on the broader impacts of our method. Our method facilitates assisting VLM to achieve better answers through knowledge retrieval. Importantly, any potentially harmful information encountered during the retrieval stage can be mitigated by appropriately restricting the scope of the knowledge base.



\end{document}